# A Two-Stage Architecture for NDA Analysis: LLM-based Segmentation and Transformer-based Clause Classification

**Ana Begnini[1], Matheus Vicente[1], Leonardo Souza[1]**

[1]Departamento de Computação - Instituto de Pesquisas Eldorado
Campinas – SP – Brazil

{`ana.begnini, matheus.rosado, leonardo.souza`}@eldorado.org.br

***Abstract.*** *In business-to-business relations, it is common to establish Non-Disclosure Agreements (NDAs). However, these documents exhibit significant variation in format, structure, and writing style, making manual analysis slow and error-prone. We propose an architecture based on LLMs to automate the segmentation and clauses classification within these contracts. We employed two models: LLaMA-3.1-8B-Instruct for NDA segmentation (clause extraction) and a fine-tuned Legal-Roberta-Large for clause classification. In the segmentation task, we achieved a ROUGE F1 of 0.95 ± 0.0036; for classification, we obtained a weighted F1 of 0.85, demonstrating the feasibility and precision of the approach.*

## 1. Introduction

Large Language Models (LLMs) are transforming legal workflows through fluent document generation and advanced contextual analysis [Taecharungroj 2023]. In the legal domain, these models are being employed in tasks such as document analysis, legal case information retrieval, and topic modeling. However, the inherent complexity of legal language, marked by domain-specific terms, ambiguity, and variation in structure, continues to pose significant challenges for natural language processing systems [Siino et al. 2025].

One practical scenario where these challenges are especially evident is the manual review of Non-Disclosure Agreements (NDAs). Legal teams are often required to analyze NDAs submitted by various external parties, each with different clause[1] structures and writing styles. This lack of standardization complicates the use of rule-based systems, while the increasing volume and urgency of contract reviews raise the risk of human error and overlooked legal inconsistencies.

To address this, we propose a two-stage architecture focused on automating key tasks in NDA analysis, namely NDA segmentation and classification. By combining a prompt-engineered large language model for NDA segmentation with a fine-tuned legal BERT classifier for clause labeling, the system reduces manual workload, ensures consistency in contract validation, and enhances legal reliability. While the current implementation targets segmentation and classification, the architecture is designed to be extensible, allowing future integration of clause correction and revision capabilities to further streamline legal review.

---

[1]In the legal context, a clause, which may comprise one or more sentences, refers to a declaration of intent that specifies the terms of an agreement, regulating or limiting the conditions established between the parties [Silva 2017].

## 2. Related Work

In this section, we organize related work according to the structure proposed by [Greco 2023], which divides language model applications in the legal domain into three main categories: Legal Domain Retrieval, Legal Document Review, and Legal Domain Prediction. These categories, in the same order as described, generally represent the execution pipeline of broader applications, particularly those implemented in corporate environments.

### 2.1. Legal Domain Retrieval

Legal domain retrieval involves systems that accurately locate legal information based on user queries. This is useful for tasks such as legal research, legal advice, and case preparation [Hambarde and Proença 2021, Mahari 2021]. Term-based techniques, such as TF-IDF and BM25, are traditionally used, including in competitions such as COLIEE (Competition on Legal Information Extraction and Entailment) [J. Rabelo and Satoh 2021]. However, more recent embedding-based approaches, such as BERT and its variations LamBERTa and LEGAL-BERT, have shown better performance [Y. Shao and Ma 2020, Almuslim and Inkpen 2020, Tagarelli and Simeri 2022]. There are also proposals that combine embeddings with TF-IDF representations, such as in the work of [D. Mamakas and Chalkidis 2022], where a modified version of LEGAL-BERT outperformed the original model, indicating more efficient and up-to-date solutions for retrieving legal documents.

### 2.2. Legal Document Review

Review of legal documents using Named Entity Recognition (NER) techniques has been essential to transform unstructured legal text into organized and searchable information [S. Shaghaghian 2020], with the support of specific datasets in the legal domain [E. Leitner 2020]. In estimating similarity between legal passages, data such as CaseHOLD [L. Zheng 2021] show that pre-training with legal vocabulary can significantly improve performance, surpassing BERT-base. Representations such as TF-IDF, Doc2Vec, and GloVe remain relevant, with TF-IDF performing notably well on smaller datasets [Almuslim and Inkpen 2020], while hybrid architectures such as Hier-SPCNet achieve high performance by integrating statutes and laws into graphs [P. Bhattacharya and Ghosh 2022]. In legal automation, models such as SPAN NLI BERT classify clauses by implication or contradiction [Y. Koreeda 2021], and ALeaseBERT is applied to identify entities and 'red flags' in contracts [S. Leivaditi 2022].

### 2.3. Legal Domain Prediction

Legal domain prediction aims to train models to identify relationships between legal passages, such as sentence coherence, which is useful for detecting argumentative gaps and predicting judicial decisions [H. Zhong and Sun 2018]. Models using Next Sentence Prediction (NSP) capture semantic dependencies in legal language, and their performance improves with pre-training on domain-specific data. BureauBERTo, adapted with texts from the public and banking sectors, achieved significant improvements [S. Auriemma and Lenci 2023]. ITALIAN-LEGAL-BERT outperformed general-purpose models in civil and criminal cases [Licari and Comandè 2024]. ConfliBERT, trained from scratch with its own vocabulary, showed improvements even without explicit

NSP [Y. Hu and D'Orazio 2018]. These studies reinforce the importance of specialized legal datasets to enhance textual prediction.

## 3. Methodology

This section presents the two-stage architecture designed to segment and classify clauses in Non-Disclosure Agreements (NDAs). The flow of this architecture was implemented using LangGraph [LangChain 2025], enabling clear definition and control of each stage. Figure 1 illustrates the proposed architecture, which is composed of two main stages.

1. **Segmenter Component**: responsible for receiving a complete NDA and split it into individual clauses;
2. **Classifier Component**: classifies each identified clause, assigning it to the most likely category based on its content.

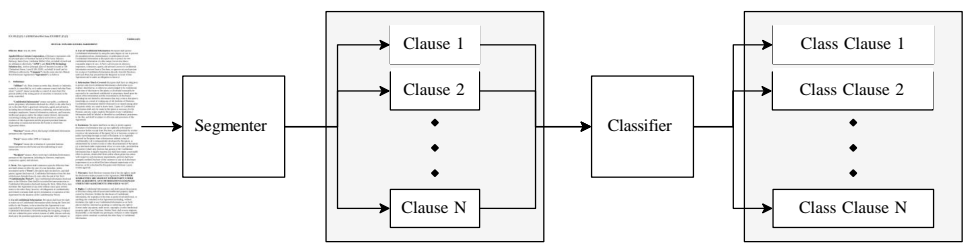

**Figure 1. Two-stage architecture for NDA analysis: segmenter and classifier.**

### 3.1. Dataset

For the experiments, documents from the public dataset **Kleister-NDA**[2] were used. This dataset contains 726 NDAs from foreign companies. All documents are written in English, available in PDF format, and are directly readable. Kleister-NDA exhibits significant diversity in writing style and textual structure, as different companies draft their NDAs with varying clauses and terminology. Additionally, the documents employ different template formats, including both single and double column layouts, which introduces further heterogeneity. This makes the dataset more representative of real-world scenarios and contributes to the robustness and generalizability of the evaluated methods.

For the experiments, a total of **322 NDA documents** were annotated, amounting to **3714 clauses**. The annotation process was carried out in two main steps: clause segmentation and classification of each clause according to a predefined set of classes. NDAs typically contain the following categories: **(1) Party Identification** indicates the involved entities; **(2) Purpose** defines whether the agreement covers a specific project or all information exchanges; **(3) NDA Type (Unilateral/Bilateral)** specifies whether the confidentiality obligation is one-sided or mutual; **(4) Definition of Confidential Information** outlines the information considered confidential, preferably identified as such; **(5) Confidentiality Obligations** sets forth responsibilities for protecting the information and the duration of confidentiality; **(6) Authorized Disclosure** identifies the people and circumstances under which information may be disclosed, such as by court order; **(7) Non-Confidential Information** lists the information excluded from protection; **(8) Liability for Damages** defines the consequences in case of breach; **(9) Competition Rights** ensures the possibility of operating in the same market, provided confidentiality is maintained; **(10) Term and Termination** establishes the duration and termination conditions of the agreement; **(11) Intellectual Property** clarifies that no transfer of rights occurs between parties; **(12) Employees** regulates the hiring of employees who have had access to the information, requiring prior notice; **(13) Governing Law and Jurisdiction** determines the applicable law and competent jurisdiction; and **(14) Additional Information** includes clauses not covered by the previous categories.

---
[2]https://github.com/applicaai/kleister-nda?tab=readme-ov-file

The entire annotation and class definition process was carried out by three legal specialists. The 322 NDAs were evenly distributed among the specialists, ensuring the quality and validity of the annotated data for model evaluation purposes. The data were labeled with multiple classes per clause, characterizing a multi-label classification problem, which means that the same clause can belong to more than one class at the same time. This scenario makes the classification task particularly challenging, especially due to the dataset imbalance. The final dataset, used in the segmentation and classification processes consists of .TXT files, where each clause is represented with the following structure:

```
[INIT_CLAUSE]
20. Governing Law. All questions concerning the construction, validity
and interpretation of this Agreement will be governed by and construed
in accordance with the domestic laws of the State of Delaware, without
giving effect to any choice of law or conflict of law provision or rule
(whether of the State of Delaware or any other jurisdiction) that would
cause the application of the laws of any jurisdiction other than the
State of Delaware. It is the intent of the parties that the provisions
of this Agreement shall be interpreted to be consistent with provisions
of Section 409A of the Internal Revenue Code.
[INIT_CLASSE]13[END_CLASSE]
[END_CLAUSE]
```

### 3.2. vLLM - *Virtual LLM*

vLLM[3] is an open-source library for LLM inference and serving, developed at the Sky Computing lab at the University of California, Berkeley. Its main goal is to provide an efficient system for memory management and parallel execution of LLMs in production environments. One of its key features is the use of the PagedAttention technique [Kwon et al. 2023], which enables dynamic memory allocation, avoiding unnecessary recomputation and optimizing GPU (Graphics Processing Unit) usage. In addition, vLLM supports multiple simultaneous requests, allowing it to serve several user sessions with low latency. The library is compatible with various popular models hosted on Hugging Face[4], including models like LLaMA [Touvron et al. 2023] and Deepseek [DeepSeek-AI 2024], which facilitate its integration into large-scale inference applications. The experiments were conducted using vLLM running on a server equipped with an NVIDIA L40S (AD102GL) GPU, based on the Ada Lovelace architecture and designed for CUDA-accelerated inference workloads.

### 3.3. Large Language Models (LLMs)

LLaMA (Large Language Model Meta AI) is a family of large-scale language models developed by Meta AI. Its architecture is based on the Transformer model [Vaswani et al. 2017], which enables large-scale text processing and generation. The LLaMA family includes models of various sizes, ranging from 7 billion to 65 billion parameters. One of its main distinguishing features is the predominant use of public datasets during training, aimed at accelerating the development of large language models and improving their robustness. Additionally, LLaMA was designed to achieve high performance with lower computational costs [Touvron et al. 2023]. These characteristics have made the model widely adopted by the academic community and by projects in resource-constrained environments, making it particularly suitable for applications that require powerful, efficient, and customizable models.

---

[3]https://github.com/vllm-project/vllm
[4]https://huggingface.co/

### 3.4. BERT Models and fine-tuning

BERT (Bidirectional Encoder Representations from Transformers) is a language model that revolutionized the field of Natural Language Processing (NLP) by introducing a bidirectional pre-training approach based on transformers. It was designed to pre-train deep bidirectional representations from unlabeled text, considering both the left and right context of each word in all layers. BERT features two pre-training objectives: Masked Language Modeling (MLM), in which some words in the text are masked and the model must predict their values based on the surrounding context; and Next Sentence Prediction (NSP), where the model receives pairs of sentences and must predict whether the second sentence is a logical continuation of the first [Devlin et al. 2019].

BERT can be fine-tuned for specific tasks. As a result, several BERT variations have emerged, such as BERTimbau [Souza et al. 2020], a version of BERT trained specifically for Portuguese, and Legal-BERT [Chalkidis et al. 2020], which was developed for tasks in the legal domain. The BERT model used in this work is **legal-roberta-base**[5] [Chalkidis et al. 2023], employed in the **classification stage**. This model is a legal-domain-oriented variant of the RoBERTa base [Liu et al. 2019]. So, for the training process of the classification model, the fine-tuning technique was used. This technique consists of adapting a model trained on large volumes of generic data to specific tasks, using smaller-scale labeled datasets [Howard and Ruder 2018]. This approach is widely used for text classification tasks with BERT [Shah et al. 2024, Yu et al. 2019]. Due to the extensive length of full NDA documents, the BERT model was not used in the segmentation stage, as its processing capacity is limited to short inputs. In contrast, LLMs offer greater capability in understanding and handling long contexts, making them more suitable for segmentation tasks.

### 3.5. Segmenter Component

The Segmenter Component represents the initial stage of the proposed two-stage system and has as its main function the decomposition of non-disclosure agreements (NDAs) into smaller units called clauses, as illustrated in Figure 2. The segmentation of legal documents poses a significant challenge, as such texts generally lack a standardized structure. Therefore, clauses can vary greatly in vocabulary and formatting, making it difficult to accurately identify their boundaries.

For this task, the vLLM framework (Section 3.2) was used together with the **Llama-3.1-8B-Instruct**[6] model (Section 3.3) for inference, making the process faster and more efficient in terms of GPU usage. The model was instructed using 499-token prompt. Input NDAs averaged 3,401 tokens (standard deviation: 2,101 tokens), and outputs 3,213 tokens (standard deviation: 1,764 tokens). Prompt adaptations were needed to help the model handle structural variations and accurately delimit clauses. Discrepancies between input and output lengths are partly due to the LLM's difficulty reproducing elements like tables and headers.

---

[5]https://huggingface.co/lexlms/legal-roberta-base
[6]https://huggingface.co/meta-llama/Llama-3.1-8B-Instruct

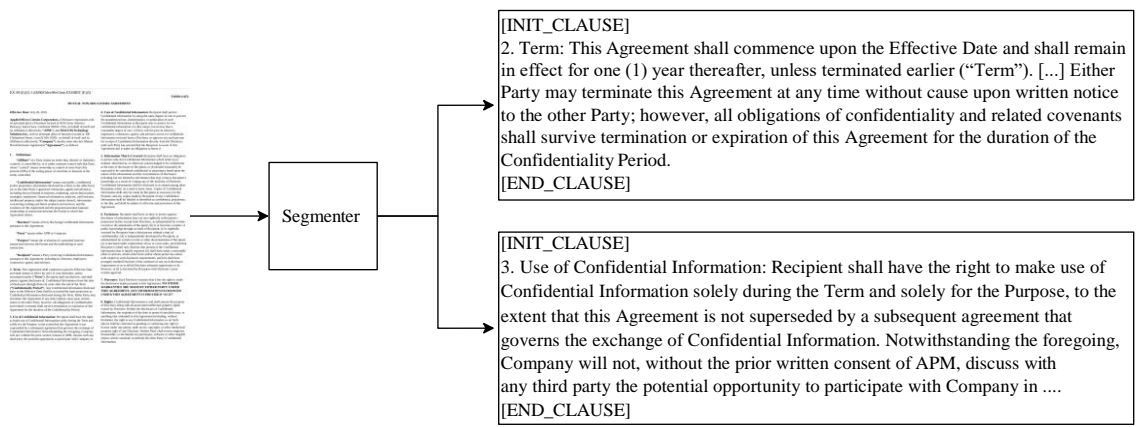

**Figure 2. Segmenter Component:** Its main task is to receive the complete NDA text and, regardless of its structure or writing style, segment it according to its clauses.

One of the main challenges in the segmentation evaluation was the absence of prior alignment between the $N$ clauses generated by the LLM and the $M$ annotated references, making direct comparison difficult. To mitigate the quadratic computational complexity ($O(N \times M)$) of an exhaustive pairwise comparison, we implemented the **Needleman-Wunsch** [Needleman and Wunsch 1970] algorithm to align predicted clauses with reference clauses. Originally developed for biological sequence alignment and later also used in textual alignment [Miller et al. 2025, Eger 2013]. This algorithm aligns two sequences by optimizing matches and penalizing insertions or deletions based on a scoring scheme. In this work, it enabled a coherent alignment between generated and annotated clauses, establishing reliable segment-level correspondence for evaluation.

The use of the Needleman-Wunsch algorithm brought a significant computational advantage. By applying a similarity threshold of 0.7 for pre-filtering, we observed an average reduction of 92.5% in the number of clause comparisons across all documents. For instance, in a case with 20 reference and 20 predicted clauses, comparisons dropped from 400 to 30. This optimization was essential for enabling the evaluation of computationally expensive metrics, such as Factual Correctness and Semantic Similarity, that rely on API calls to language models [OpenAI 2024] and incur financial cost.

For model validation, different evaluation metrics were adopted, including ROUGE, Factual Correctness, and Semantic Similarity. Each of these metrics is detailed below:

- **ROUGE** (Recall-Oriented Understudy for Gisting Evaluation): a widely used metric to assess lexical similarity between texts, based on the overlap of n-grams between the generated output and a reference text [Lin 2004]. In this work, the

ROUGE-1 variant was used, which considers unigrams, allowing the measurement of how much relevant content from the annotated clauses is preserved by the model, through recall, precision, and F1-scores.
- **Factual Correctness**: a metric that evaluates the factual accuracy of the generated response in relation to the reference, by decomposing the text into factual statements and verifying them via textual inference [Ragas Documentation 2025a]. This is especially relevant in this work to ensure that segmentation fully preserves the information contained in the original clauses, without omissions or distortions. The evaluation was performed using the GPT-4.1-Nano model [OpenAI 2023], chosen for its robustness in factual verification.
- **Semantic Similarity**: metric used to evaluate how close, in terms of meaning, the generated segment is to the reference clause. Considering the overall meaning of the texts, this metric provides valuable insights about the fidelity of the segmentation. Each clause is converted into a dense vector (embedding) and compared using cosine distance [Ragas Documentation 2025b]. In this work, the text-embedding-3-large model from OpenAI [OpenAI 2024] was used to semantically represent the segments.

### 3.6. Classifier Component

The Classifier Component represents the second stage of the proposed two-stage system and is responsible for assigning a specific semantic category to each previously segmented clause. The goal is to identify, based on the textual content of each clause, its respective legal class, as defined in the annotation scheme described in Section 3.1.

For classification, fine-tuning was performed on the **legal-roberta-base** model [Chalkidis et al. 2023] (Section 3.4), chosen for its ability to process long textual sequences and its relatively small number of parameters, which is suitable for the dataset size. The dataset used consists of 3,714 clauses, each of which may be associated with multiple classes, characterizing the problem as a multi-label classification. As a result, a significant class imbalance was observed: Class 14 (Section 3.1) accounts for approximately 48.9% of the occurrences, followed by Class 5 with around 17.2%, while the remaining 33.9% are distributed among the other classes.

The training process was configured with 3 epochs, a learning rate of 1e-5, weight decay of 0.01, warmup ratio of 0.1, and no dropout (dropout = 0.0). To address the class imbalance, the Focal Loss [Lin et al. 2018] function was employed, with parameters $a$ = 0.25 and $\gamma$ = 2. Moreover, for model validation, different evaluation metrics were adopted, including F1-Macro, F1-Weighted, Hamming Loss, and MCC (Matthews Correlation Coefficient). Each of these metrics is detailed below:

- **Macro F1**: calculates the average F1-score of each class individually, treating all classes with equal weight regardless of their frequency in the dataset [Van Asch 2013]. It is used in multi-label and imbalanced classification contexts, as it allows the model's performance to be evaluated fairly across both majority and minority classes [Grandini et al. 2020].
- **Weighted F1**: computes the F1-score weighted by the frequency of each class [Pedregosa et al. 2011]. Thus, more frequent classes have a greater impact on the final score, making this metric suitable for analyzing the overall model performance on imbalanced data.

- **Hamming Loss**: measures the error rate in multi-label classifications by evaluating the proportion of incorrect labels relative to the total number of possible labels [Zhang and Zhou 2006]. This metric provides a clear view of prediction accuracy at the individual label level by considering all classes equally, even in cases where clauses have two or more classes.
- **MCC**: is a statistical metric widely recognized for providing a balanced evaluation of binary classification quality, even in imbalanced class scenarios [Matthews 1975]. Adapted for multi-label problems, MCC captures the correlation between predicted and true classes.

## 4. Results

The results of the metrics obtained by the two components are presented and discussed below: the Segmenter (Section 4.1) and the Classifier (Section 4.2).

### 4.1. Segmenter Component

The results shown in Tables 1 and 2 demonstrate the performance of the proposed segmentation model, both at the document level and at the segment level, respectively.

A total of 314 NDAs containing 3,714 clauses were used. Confidence intervals of 95% were applied to express the uncertainty associated with the ROUGE, Factual Correctness, and Semantic Similarity metrics. To compute these intervals, the Student's t-distribution was applied, as the estimates were obtained from a finite and small sample with unknown population variance. The t-distribution is particularly suitable in this context, as it adjusts the interval width according to the sample size, incorporating the additional uncertainty stemming from estimating the standard deviation from the data itself. Thus, using the Student's t-distribution ensures that the intervals obtained represent, with 95% confidence, the range within which the true population mean of the evaluated metrics is expected to fall, providing a more rigorous and grounded statistical analysis.

The comparison between the generated documents to the references globally evaluates how much of the content from the reference NDA was effectively preserved in the text segmented by the LLM (see Table 1). The ROUGE results demonstrate the model's robustness in the task of NDA contract segmentation. The ROUGE-Precision value of 0.99 indicates that the content segmented by the LLM is highly faithful to the expected vocabulary and structure in the reference clauses. Meanwhile, the ROUGE-Recall value, with an average of 0.91, reinforces that most of the information present in the reference documents was preserved in the segmented outputs. This metric is particularly relevant for assessing the model's ability to retain the essential content of the source text, ensuring that the segmentation does not compromise the informational completeness of the document. These results indicate an efficient and consistent segmentation at the document level.

The segmentation quality was assessed by the direct correspondence between the LLM-predicted clauses and the reference ones, the results are presented in Table 2. To ensure a focused analysis on relevant pairs, we established a filtering criterion: only pairs with a Needleman-Wunsch alignment higher than a threshold of 0.7 were considered. In the subset of samples that satisfied this condition, the average alignment reached 0.98 ± 0.0015. This high value indicates a near-perfect correspondence for the correctly mapped clauses, confirming the robustness and accuracy of the segmentation model and also providing a reliable basis for evaluating the remaining metrics. The ROUGE metric shows consistent results, with 0.98 precision and 0.94 recall, indicating that the textual structure of the clauses was maintained. The Factual Correctness yielded an average of 0.95, indicating that the essential information of each clause was correctly preserved. Finally, the Semantic Similarity, at 0.98, reinforces that the generated segments remain highly aligned with the meaning of the reference clauses. These results indicate that the model not only correctly identifies clause boundaries but also preserves their factual and semantic integrity.

**Table 1. Average metrics comparing Reference Document vs. Generated Document and their confidence intervals.**

|  | Metric Average ± Confidence Interval |
|---|---|
| ROUGE-Recall | 0.91 ± 0.0112 |
| ROUGE-Precision | 0.99 ± 0.0019 |
| ROUGE-F1-Score | 0.94 ± 0.0084 |

**Table 2. Average metrics comparing Reference Segment vs. Generated Segment and their confidence intervals.**

|  | Metric Average ± Confidence Interval |
|---|---|
| ROUGE-Recall | 0.94 ± 0.005 |
| ROUGE-Precision | 0.98 ± 0.0019 |
| ROUGE-F1-Score | 0.95 ± 0.0036 |
| Factual Correctness | 0.95 ± 0.0044 |
| Semantic Similarity | 0.98 ± 0.0027 |

### 4.2. Classifier Component

From the dataset of 3,714 clauses, a stratified sample was created for multi-label classification and split into three subsets: 2,676 clauses (80% of the dataset) for training, 298 for validation (10% of the training set), and 740 for testing (20% of the dataset). Table 3 presents the classification model's performance on the validation and test sets.

**Table 3. Classification Model Metrics**

|  | Loss | macro F1 | weighted F1 | Hamming Loss | MCC |
|---|---|---|---|---|---|
| Validation | 0.01 | 0.67 | 0.84 | 0.03 | 0.82 |
| Test | — | 0.69 | 0.85 | 0.03 | 0.84 |

We observe that the macro F1 metric presented relatively low values in our multi-label classification problem, with results of 0.67 in validation and 0.69 in testing. This difference is primarily due to the model's inability to generalize well to underrepresented classes in the dataset. Specifically, four classes were identified with fewer than 100 samples each, and this scarcity negatively impacted the macro F1 performance, which assigns equal weight to all classes regardless of their frequency. In contrast, the weighted F1 metric, which weighs the performance of each class by its relative frequency, showed significantly higher values: 0.84 in validation and 0.85 in testing. This contrast highlights the impact of class imbalance, indicating that the model was more effective in learning patterns from the majority classes but had limitations in recognizing the minority classes.

The Hamming Loss and MCC metrics also reflect the challenges posed by the multi-label and imbalanced nature of the problem. Hamming Loss showed low values, 0.03 in validation and 0.03 in testing, indicating that, on average, the model makes few errors per predicted label. However, since this is an average over all labels, good results on the majority classes may smooth out the impact of errors on minority classes. Additionally, the MCC values of 0.82 in validation and 0.84 in testing point to a strong overall correlation between predictions and true labels. These results indicate that the model maintains good general predictive quality, being effective not only for the most frequent classes but also showing stable performance across the entire dataset.

## 5. Conclusion and Future Work

In this work, a two-stage architecture was proposed for the analysis of Non-Disclosure Agreements (NDAs), focusing on clause segmentation and classification using large language models (LLMs) integrated with the vLLM framework and specialized BERT models for legal document classification.

The segmentation process was carried out using the Llama-3.1-8B-Instruct model integrated with vLLM, enabling efficient handling of lengthy documents. Prompt engineering guided the model to produce high-quality segmentation, as confirmed by metrics such as ROUGE, Factual Correctness, and Semantic Similarity. The Needleman-Wunsch algorithm was employed during evaluation to align predicted and reference clauses, ensuring accurate comparison. For classification, a fine-tuned legal-roberta-base model was used, addressing the dataset's multi-label and imbalanced nature. Techniques like Focal Loss and stratified splitting enabled robust evaluation with metrics including macro F1, weighted F1, Hamming Loss, and MCC, demonstrating the model's effectiveness in capturing frequent classes and maintaining strong agreement with reference annotations.

A significant challenge faced during this study was the acquisition of a suitable dataset. Due to the confidential nature of NDAs, companies generally do not release such documents, making data collection particularly difficult. This scarcity of annotated NDAs limited the number of samples available, especially for minority classes, which in turn impacted the model's ability to generalize across all categories. Despite these constraints, the results obtained were promising and laid a strong foundation for future improvements.

For future work, we intend to improve the classifier's performance through techniques such as data augmentation, paraphrase generation with generative models, and additional strategies beyond Focal Loss, aiming to mitigate the imbalance across the 14 classes. Furthermore, the segmentation and classification stages will be integrated into a complete system to automate the review and correction of Non-Disclosure Agreements (NDAs). This system will include specialist agents responsible for analyzing specific clauses, identifying potential inconsistencies or non-compliances, and suggesting improvements.